\documentclass{article}

\usepackage{PRIMEarxiv}

\usepackage[utf8]{inputenc} 
\usepackage[T1]{fontenc}    
\usepackage{hyperref}       
\usepackage{url}            
\usepackage{booktabs}       
\usepackage{amsfonts}       
\usepackage{nicefrac}       
\usepackage{microtype}      
\usepackage{lipsum}
\usepackage{fancyhdr}       
\usepackage{graphicx}       
\graphicspath{{media/}}     
\usepackage{hyperref}

\usepackage{algorithm, algorithmic}
\usepackage{amsmath}

\title{NeRF synthesis with shading guidance
}

\author{
  Chenbin Li, Yu Xin, Gaoyi Liu, Xiang Zeng, Ligang Liu \\
  University of Science and Technology of China \\
  Hefei\\
  China\\
  \texttt{\{lichenbin, xy0731, Zengx56, liugaoyi\}@mail.ustc.edu.cn} \\
  \texttt{lgliu@ustc.edu.cn} \\
}

\begin{document}
\maketitle

\begin{abstract}
	The emerging Neural Radiance Field (NeRF) shows great potential in representing 3D scenes, which can render photo-realistic images from novel view with only sparse views given. 
	However, utilizing NeRF to reconstruct real-world scenes requires
	images from different viewpoints, which limits its practical application. This problem can be even more pronounced for large scenes.
	In this paper, we introduce a new task called NeRF synthesis that utilizes the structural content of a NeRF patch exemplar to construct a new radiance field of large size.
	We propose a two-phase method for synthesizing new scenes that are continuous in geometry and appearance. 
	We also propose a boundary constraint method to synthesize scenes of arbitrary size without artifacts. 
	Specifically, we control the lighting effects of synthesized scenes using shading guidance instead of decoupling the scene. 
	We have demonstrated that our method can generate high-quality results with consistent geometry and appearance, even for scenes with complex lighting.
	We can also synthesize new scenes on curved surface with arbitrary lighting effects, which enhances the practicality of our proposed NeRF synthesis approach.
\end{abstract}


\section{Introduction}

Physically-based rendering (PBR) plays a critical role in a wide range of computer graphics applications, including virtual reality, augmented reality, and video games. However, it struggles to accurately represent real-world scenes that involve complex geometry and textures, such as fine hair and grass. Recently, NeRF~\cite{mildenhall2021nerf} demonstrated a state-of-the-art approach to novel view synthesis using an implicit volumetric field that outputs density and radiance values. By training the field on a sparse set of input views, NeRF can efficiently generate photo-realistic images from any novel view using a ray-marching technique.
Recent research has shown that NeRF is capable of high-fidelity modeling of complex natural scenes~\cite{gao2022nerf} and can be applied to large-scale scene modeling~\cite{tancik2022block, hong2022headnerf}.

While NeRF is adept at handling complex scenes, capturing multiple views for large-scale scenes can present challenges due to the large number of views that need to be captured. Although recent efforts have been made to improve the efficiency of NeRF training~\cite{Chen2022ECCV, mueller2022instant}, it remains time- and memory-intensive for large-scale scenes. The question arises: is it truly necessary to capture all real-world objects in a scene through cameras and model them by NeRFs in order to create virtual models of real scenes? For objects such as fur, lawns, and clouds, which possess clear repetitive structures. Modeling them on a large scale can be tedious and laborious. We observe that these objects are isotropic in the plane and are well-suited for block-by-block generation. Inspired by texture synthesis~\cite{raad2018survey}, we propose NeRF synthesis, which is the process of algorithmically constructing a large radiance field from a smaller exemplar by making use of its structural content. 

This task presents several challenges. Firstly, unlike 2D images, NeRF is a function of viewpoint, which requires the synthesized results are rendered consistently from every viewpoint. Secondly, NeRF couples lighting information with material  information, posing a significant challenge in ensuring plausible lighting in the synthesized results. A common method to achieve plausible lighting is to completely decouple the scene~\cite{zhang2021nerfactor, rudnev2022nerf, yang2022ps}, obtaining information such as geometry, material, and lighting, and then using traditional rendering pipelines to relight the scene. However, decoupling scenes with complex geometric details, particularly those with grass, fur, etc., poses a significant challenge. Similary, BTF~\cite{tong2002synthesis}  depends on lighting and viewpoints. In contrast to NeRF, the rendering result of BTF from different viewpoints and lighting conditions can be aligned to the same 2D image, enabling the extraction of features that incorporate both the viewpoint and lighting. However, since NeRF uses volume rendering to obtain images from different viewpoints, these images do not have a corresponding relationship at the same position, making it impossible to directly apply BTF synthesis method to NeRF synthesis.

To address the aforementioned issues, first, we utilize shading vector to control the lighting of the synthesized scene. The shading vector refers to the shading value of a voxel column in the radiance field at $M$ different viewpoints. By comparing the distance between two shading vectors, we can quantify their lighting differences. All shading vectors in a scene together form a shading map. 
To relight the scene, the shading vector of the exemplar is obtained by intrinsic decomposeition-based method, and then a ray-tracing-based method is proposed to convert lighting information into a shading map guider. With the aid of the shading map and the shading map guider, we can guide the synthesis process by selecting patches with the correct lighting information.
Second, to ensure both geometric and appearance continuity in the synthesized results, especially when dealing with complex texture and geometry, we propose a two-phase method. For the boundaries of the scene, we introduce a boundary-constrained synthesis method to prevent artifacts at the boundaries when synthesizing large scenes.

To summarize, this paper makes the following contributions:

\begin{itemize}
	\item Proposing a two-phase method to ensure both geometry and appearance continuity, and a boundary-constrained synthesis approach to avoid artifacts resulting from obscured	parts of the radiance field.
	\item Proposing a shading-guided synthesis method to synthesize large-scale scene with reasonable lighting. This method allows for relighting scenes without the need for decoupling the scene.
\end{itemize}

\section{Related work}

\subsection{Planar texture synthesis}
The texture is a kind of pattern with repetition, such repetition does not need to be spatially periodic, but there is some similarity between any two regions. This special repetition makes it possible to synthesize large-scale textures from a local texture sample.

Methods of texture synthesis can be sorted into two categories: parametric methods and non-parametric methods. Parametric methods treat texture as a parametric math model. By adjusting the pre-defined parameters of a pure-noise image and matching them with those of the input texture sample, an output texture with similar parameters to the input texture will be produced. In many cases, these parameters are related to some statistics of features of the input texture, such as histograms of frequency bands, wavelet coefficients, or responses to a set of filters~\cite{Heeger1995, DoBonet1997, Portilla2000}.

Correspondingly, non-parametric methods no longer care about those statistics of texture and establish some rules for directly constructing the output texture. One popular approach is modeling texture as a Markov Random Field (MRF)~\cite{Efros1999}, which assumes that local patterns of texture have a stationary distribution. This allows us to infer the pattern in blank areas of output according to their synthesized neighborhood and induces a family of schematic methods called exemplar-based methods. These methods move one part from the input texture to output texture per step and try to maintain the stationary distribution for every move.

One branch of the exemplar-based method is the pixel-based method, which moves one pixel per step~\cite{Efros1999, wei2000fast, Ashikhmin2001, Kwatra2005}. The pixel from the input texture whose neighborhood has a minimum difference from that of the on-processing pixel from the output texture will be chosen. Another branch of the exemplar-based method is the patch-based method~\cite{Wu2004, Efros2001, Kwatra2003, Rafi2019}, which moves a patch from the examplar per step according to neighborhood-comparing criteria similar to that adopted by pixel-based method. Patch-based methods can be generally more efficient than pixel-based methods, and perform better in preserving large-scale features.

While the above works performed neighborhood comparisons on color channel information, Tong et al. \cite{Tong2002} introduced a concept of `texton' that integrates multi-orientation-lighting information together with responses of filters. By searching neighborhoods in texton space instead of pixel space, they successfully synthesized consistent BTF textures for arbitrary manifold surfaces by the exemplar-based method. This inspires the notion that traditional methods possess the capability to solve high-dimensional texture synthesis.

\subsection{Geometric texture synthesis}
The concept of texture also applies to three-dimensional (3D) objects. 3D objects with repetitive geometric patterns can be spotted everywhere. From the surfaces of natural objects like vegetation, terrain, waves, and fur, or from artificial objects, such as knitting, bricks wall, and buildings. Textures like fur and hair consist of rather complex surfaces. The volume integrates their average surface properties and color information that should be a suitable representation for this kind of textures~\cite{Kajiya1989, Neyret1998}. Other textures like terrain and wave have determined boundaries, which can be precisely represented by meshes. Zhou et al. \cite{Zhou2006} extended the patch-based method to handle mesh texture synthesis tasks and employed mesh deformation techniques to blend overlapping areas. Their results can be seamlessly applied to manifold surfaces, resulting in objects with intricate surface geometries.

\subsection{NeRF-based generative models}
Due to the similarity between NeRF and images in representation, successful generation techniques on images can also be applied to NeRF. For example, one can directly treat the radiance field as the parameter to optimize and define an adversarial loss based on images obtained through volumetric rendering and real images to train a generative adversirial network (GAN)~\cite{GAN2014}.
This enables the learning of GAN-based NeRF generation models~\cite{GIRAFFE2020, GRAF2020, piGAN2020}.
Alternatively, one can perform denoising of a probabilistic diffusion process applied to 3D radiance fields to learn a diffusion-based NeRF generation models in 3D space~\cite{muller2022diffrf,shue20223d}.
However, these methods differ from NeRF synthesis in several key aspects. Firstly, these methods generate objects within the same distribution as the original data, limiting their ability to create objects of varying sizes. Secondly, these generative methods often require large amounts of training data, whereas NeRF synthesis only requires a single exemplar.
While current works such as \cite{wang2022singrav,weiyu23Sin3DGen} focus on scene generation from a single example represented by NeRF, our work focuses on synthesizing tiled scenes such as grass and carpets, achieving continuity in geometry and texture while maintaining control over lighting.

\section{Preliminaries}

\begin{figure*}
	\centering
	\includegraphics[width=\linewidth]{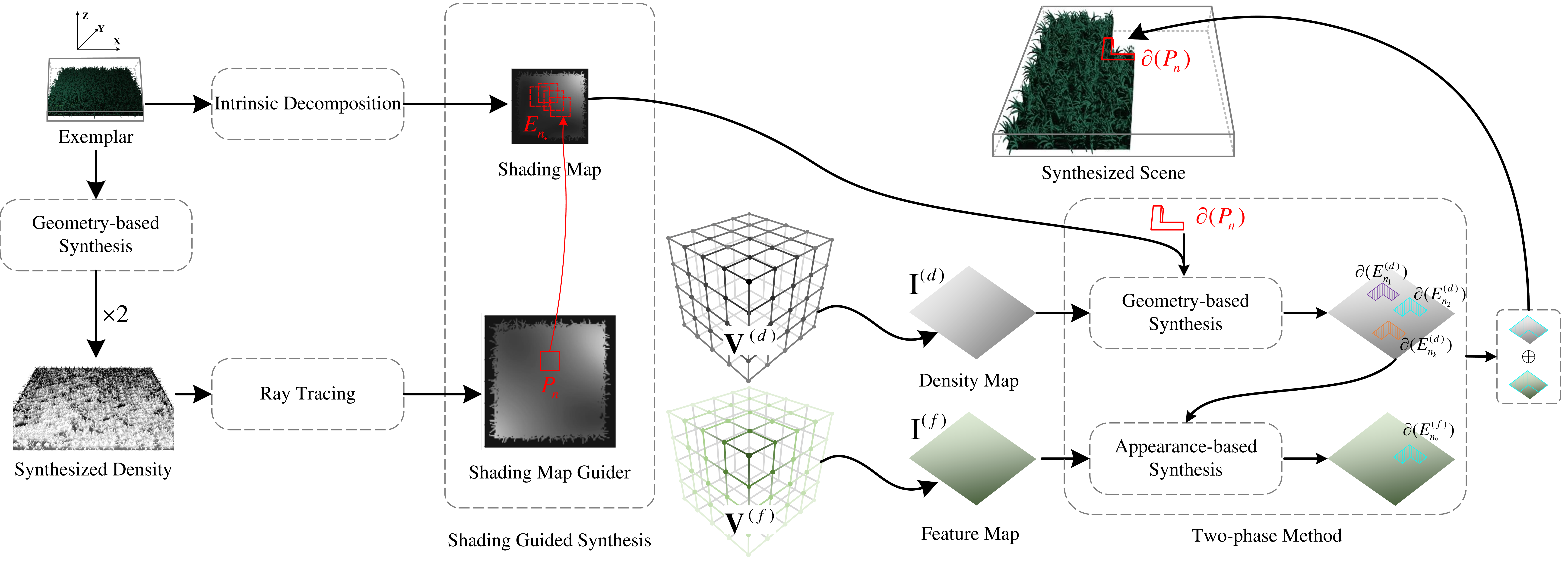}
	\caption{Pipeline of our proposed method. A density field ($\mathbf{V}^{(d)}$) and a color feature field ($\mathbf{V}^{(f)}$) are provided to represent the exemplar. The shading map extracted from the exemplar and shading map guider are also provided for synthesis with lighting constrained. For the patch $P_n$ to be synthesized, shading guided synthesis first finds the closest $k_s$ patches by the shading distance between $P_n$ and patches in the exemplar. Then, the two-phase method is applied to these patches to filter the patches with high response values in terms of density and color. In the first phase, $k_g$ candidate patches with geometric continuity are found based on $\mathbf{V}^{(d)}$. In the second phase, one patch is selected for color continuity in the overlap area based on $\mathbf{V}^{(f)}$. The found patch's density field and radiance field are stitched onto the target area to complete one step of NeRF synthesis.
		\label{pipeline}}
\end{figure*}

\subsection{Direct voxel grid optimization}

Direct voxel grid optimization (DVGO)~\cite{sun2022direct} is an explicit-implicit hybrid representation, which uses a voxel-grid model to represent a 3D scene for novel view synthesis. As shown in Fig.~\ref{pipeline}, the implicit modalities are stored explicitly in two voxel grids: a density grid $\mathbf{V}^{(d)}$ and a color feature grid $\mathbf{V}^{(f)}$.
For a pixel-rendering ray $\boldsymbol{r}$, DVGO samples $n$ points between the pre-defined near and far planes and calculates the feature of each sampled point $x_i$ using trilinear interpolation:

\begin{equation}
	interp(x_i, \mathbf{V}):(\mathbb{R}^{3}, \mathbb{R}^{C \times N_x \times N_y \times N_z}) \rightarrow \mathbb{R}^{C}.
\end{equation}

DVGO uses the shifted softplus mentioned in Mip-NeRF \cite{barron2021mip} as density activation ($i.e.$, a mapping from $\mathbb{R}$ to $\mathbb{R}_{\geq 0}$):

\begin{equation}
	\sigma_i = softplus(\ddot{\sigma}_i) = log(1 + exp(\ddot{\sigma}_i + b)),
\end{equation}
where $\ddot{\sigma}_i$ represents the value stored in $\mathbf{V}^{(d)}$, and the shift $b$ is a hyperparameter. The color feature will be mapped to a view-dependent color emission $\textbf{\emph c}_i$ by a shallow multilayer neural network parameterized by $\Theta$:

\begin{equation}
	\textbf{\emph c}_i = MLP_{\Theta}^{(rgb)}(interp(x_i, \mathbf{V}^{(f)}), x_i, d),
\end{equation}
where $d$ represents the direction of the ray $\textbf{\emph r}$.
Positional encoding is applied to encode $x_i$ and $d$ for the $MLP_{\Theta}^{(rgb)}$.
Finally, the $n$ results $\{\sigma_i, \textbf{\emph c}_i\}_{i=1}^n$ are accumulated into a single color $\hat C(\textbf{\emph r})$ using the volume rendering quadrature:

\begin{equation}
	\hat C(\textbf{\emph r}) = (\sum_{i=1}^n T_i \alpha_i \textbf{\emph c}_i) + T_{n + 1} \textbf{\emph c}_{bg},
\end{equation}

\begin{equation}
	\alpha_i = alpha(\sigma_i, \delta_i) = 1 - exp(-\sigma_i \delta_i),
\end{equation}

\begin{equation}
	T_i = \prod_{j=1}^{i-1} 1 - \alpha_j,
\end{equation}
where $\alpha_i$ represents the probability of termination at the point $x_i$; $T_i$ denotes the accumulated transmittance from the near plane to point $x_i$; $\delta_i$ indicates the distance to the adjacent sampled point, and $\textbf{\emph c}_{bg}$ represents a pre-defined background color.
In addition, DVGO shows that the post-activation for density:
$
\alpha^{(post)} = alpha(softplus(interp(x, \mathbf{V}^{(d)}))),
$
i.e. applying all the non-linear activation functions after the trilinear interpolation, is capable of producing sharp surfaces, thus requiring fewer cells for surface detail.

\subsection{Patch-based texture synthesis}\label{Patch-Based}

\begin{figure}
	\centering
	\includegraphics[width=0.6\textwidth]{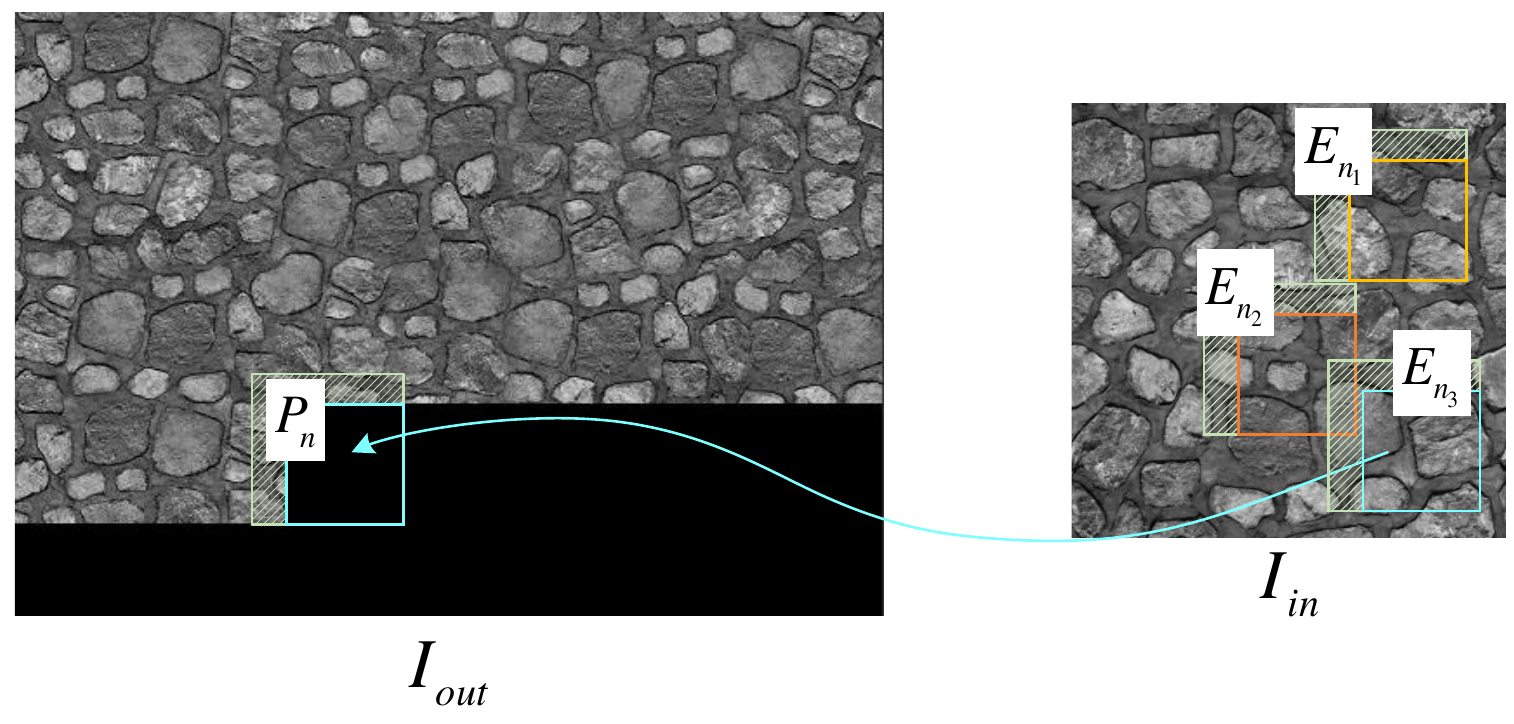}
	\caption{$I_{in}$ is an example texture image. A new, larger texture image $I_{out}$ is waiting to be synthesized based on $I_{in}$. A patch $P_n$ to be synthesized is match with $E_{n_i}$ $(i=1,2,3)$, and $E_{n_3}$ is selected to synthesize the texture patch $P_n$. It should be emphasized that patches in various locations may have different types of overlap. This figure only illustrates the most common case.
		\label{synthesis}}
\end{figure}

As shown in Fig.~\ref{synthesis}, to synthesize a new texture patch $P_n$ of new texture image $I_{out}$ from the input texture image $I_{in}$, the patch-based algorithm pastes an example patch $E_n$ of the input sample texture $I_{in}$ into the synthesized texture $I_{out}$ step by step. The selection of $E_n$ is based on the patches already pasted in $I_{out}$ to avoid mismatching features across patch boundaries. Specifically, we say that $P_{n}$ and $E_{n}$ match if 

\begin{equation}
	d(\partial(P_{n}), \partial(E_{n})) < \eta,    
\end{equation}
where $\partial(\cdot)$ represents the boundary zone of a texture patch, and $d(\cdot,\cdot)$ represents the L2 distance between two texture patches. Here, $\eta$ is a prescribed constant, and the overlap size is a pre-defined constant.
All the matched example patches are directly searched in all example patches in $I_{in}$ that have the known $\partial(P_{n})$ as their boundary zones.

Liang et al.~\cite{liang2001real} firstly used the overlap region to initial optimized kd-trees for accelerating the approximate nearest neighbors (ANN) search. In each step, the ANN search is used to find the $k$ most similar patches $\{E_{n_i}\}_{i=1}^k$.
Then, a probability density function (PDF) is computed based on $d(\partial(P_{n}), \partial(E_{n_i}))$ for $i=1$ to $k$, and a patch is randomly chosen according to the computed PDF.
Finally, the chosen patch is pasted into the synthesized texture $I_{out}$, and the overlap area is blended to provide a smooth transition between adjacent texture patches.

\section{Method}
NeRF uses an implicit method to represent the radiance field by an MLP, which is not suitable be direct used in patch-based synthesis. 
Therefore, we utilize DVGO, which is an explicit-implicit hybrid representation. 
Since our method is not limited to DVGO, we continue to refer to it as NeRF synthesis.
For both the density grid $\mathbf{V}^{(d)}$ and the feature grid $\mathbf{V}^{(f)}$ of DVGO, we consider them as 2D grids. 
This is because only two dimensions exhibit clear repeatability, even we want to synthesize a three-dimensional object, as shown in Fig~\ref{pipeline}.
Taking grass as an example, it would be inefficient and unnecessary to synthesize the roots of the grass first and then search for matching leaves. Treating grass as a unified entity would be more reasonable. 
Therefore, we flatten the 3D grid into a 2D grid along the $\mathbf{Z}$-axis, resulting in $\mathbf{I}^{(d)}$ and $\mathbf{I}^{(f)}$. Consequently, we can directly employ 2D texture synthesis methods for 3D NeRF synthesis. The only distinction is that, in NeRF synthesis, each unit stores density and feature values instead of RGB.

\subsection{Two-phase method\label{sec:two-phase}}
Unlike 2D texture synthesis, NeRF synthesis requires ensuring both geometry and appearance continuity simultaneously. Therefore, it is necessary to consider them together during synthesis.
We directly concatenate the two grids ($\mathbf{V}^{(d)}$ and $\mathbf{V}^{(f)}$) and treat them as a combined 2D image, denoted as $\mathbf{I}^{(d \oplus f)}$. 
The distance between different patches is measured using the L2 distance.
The patch-based synthesis method is applied directly on this grid to obtain the synthesized results. We take this method as a baseline. 
However, a significant challenge arises in determining the weighting between density and color. Different scenes often require different suitable weights. 
As depicted in Fig.~\ref{two-phase}, using a higher weight for density leads to more pronounced color inconsistencies in the synthesized results, whereas employing a higher weight for color features results in more severe geometric inconsistencies. Additionally, in order to capture viewpoint-dependent color variations, $\mathbf{I}^{(d \oplus f)}$ tends to have high dimensionality, leading to increased search time during the process.

To address the aforementioned challenges, a two-phase method is proposed, as illustrated in Fig~\ref{pipeline}, including geometry-based synthesis and appearance-based synthesis.
In the first phase, for a voxel patch $P_n$ to be synthesized, the $k_g$ best density voxel patches $E_{n_i}^{(d)}(i=1,2,..., k_g)$ are selected based on the similarity of $d(\partial(P^{(d)}), \partial(E^{(d)}))$. 
The assumption is that due to their relatively high similarity in the overlapping region of the density field, these patches also exhibit relatively high geometric continuity at the position to be synthesized.
To ensure the continuity of appearance in the synthesized results, the similarity of $d(\partial(P_n^{(f)}), \partial(E_{n_i}^{(f)}))(i=1,2,..., k_g)$ is calculated for the selected $k_g$ best patches. These similarities are then utilized as the likelihood of each example patch being selected during the random selection process, ensuring randomness in the synthesis procedure.
Finally, the overlapping regions of the selected patch $E_{n_*}$ and $P_n$ are blended using inverse distance weighting.

\textit{Why prioritize density over color in the synthesis process?} The synthesis process operates at the voxel column level, where the color feature inherently incorporates geometric information to some extent. By prioritizing density first, we ensure the continuity of geometric, while subsequent considerations of appearance are less affected by density.

\subsection{Boundary-constrained synthesis method}

\begin{figure}
	\centering
	\includegraphics[width=0.8\textwidth]{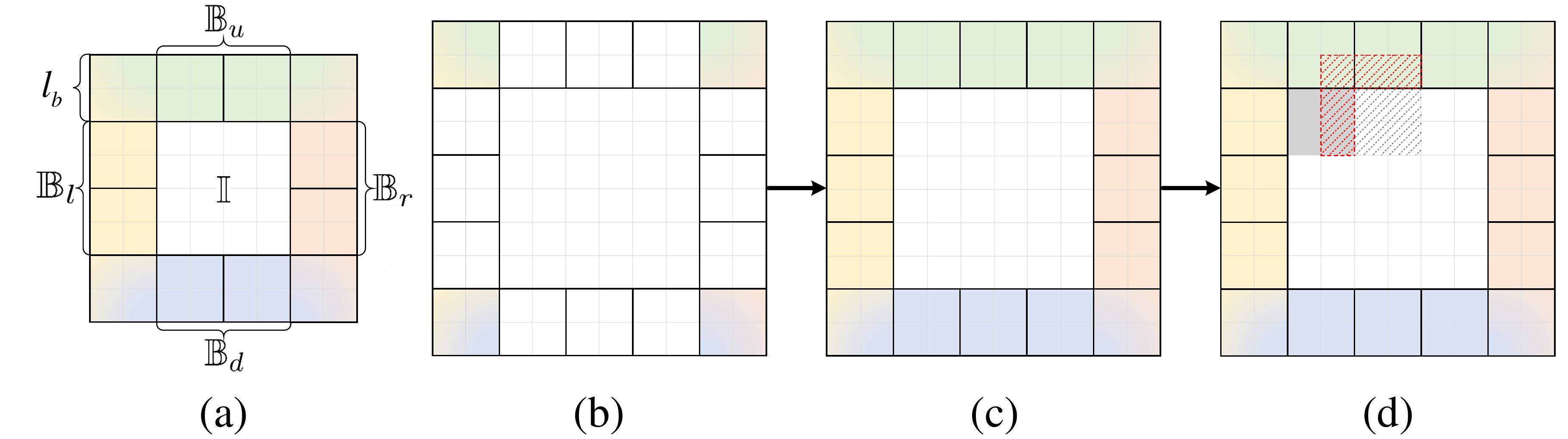}
	\caption{
		(a) Original exemplars for synthesis. 
		(b) Four corner patches are set to initial the canvas. 
		(c) The four boundaries are synthesized using the boundary-constrained synthesis method. The patches in $I_{out}$ with color come from from the corresponding patches in $I_{in}$.
		(d) The boundary serves as a constraint for further synthesizing the inner region. The gray area represents the synthesized part. The gray slash part is the area to be synthesized. The white part indicates the unsynthesized area. The red slash represents the overlap area of the regions to be synthesized.
		\label{edge_method}}
\end{figure}

As shown in Fig.~\ref{boundary}, the two-phase method faces challenges when handling synthesis on the boundary. This is because the occluded interior parts are synthesized onto the boundary. Due to the lack of pixel value supervision for these occluded parts during the DVGO training process, they exhibit artifacts. 

To address this issue, as shown in Fig.~\ref{edge_method}, we divide the entire radiance field into five parts: the upper boundary $\mathbb{B}_u$, the lower boundary $\mathbb{B}_d$, the left boundary $\mathbb{B}_l$, the right boundary $\mathbb{B}_r$, and the interior area $\mathbb{I}$. The length of each boundary is denoted as $l_b$.
Initially, we place the four corner patches from the original field directly onto the corresponding corners of the synthesized result. These corner patches are essential as they have boundaries in two directions and cannot be replaced.
Next, we synthesize each of the four boundaries in the output by utilizing the corresponding set of boundary patches from the input samples. This synthesis process can be viewed as synthesizing individual strips.
Finally, we employ the synthesized boundaries as constraints for synthesizing the interior region. We accomplish this by utilizing example patches from the interior area $\mathbb{I}$.
It is important to note that the physical boundaries of the scene may span a certain length in the field. Therefore, it is necessary to choose an appropriate length to handle the boundaries separately. For the sake of clarity, we assume that the lengths of the four boundaries are set to the same parameters.

\subsection{Shading guided NeRF synthesis}

\begin{figure}
	\centering
	\includegraphics[width=0.6\linewidth]{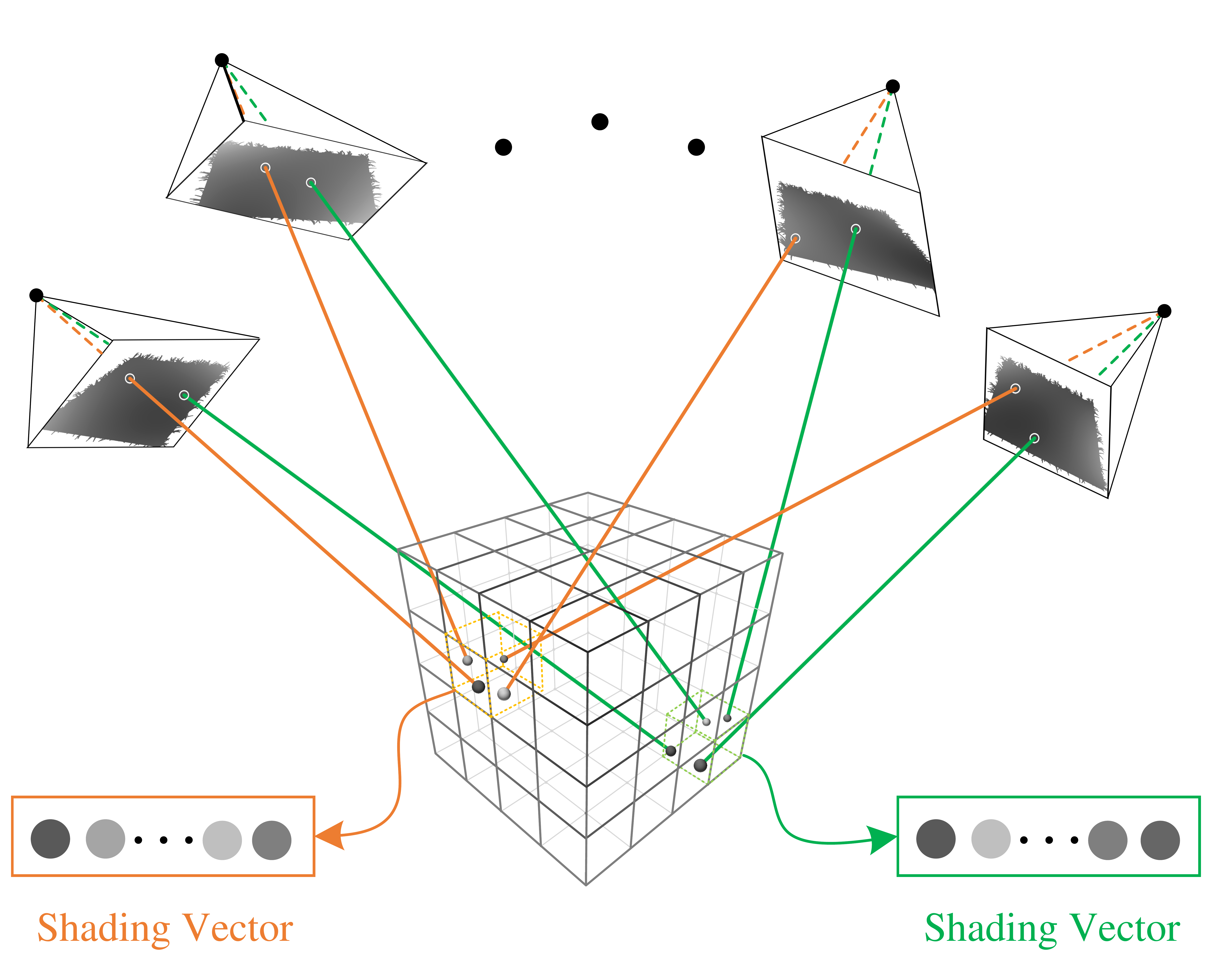}
	\caption{Shading vector from intrinsic decomposition.
		\label{shadeton_pie}}
\end{figure}

\begin{figure}
	\centering
	\includegraphics[width=0.6\linewidth]{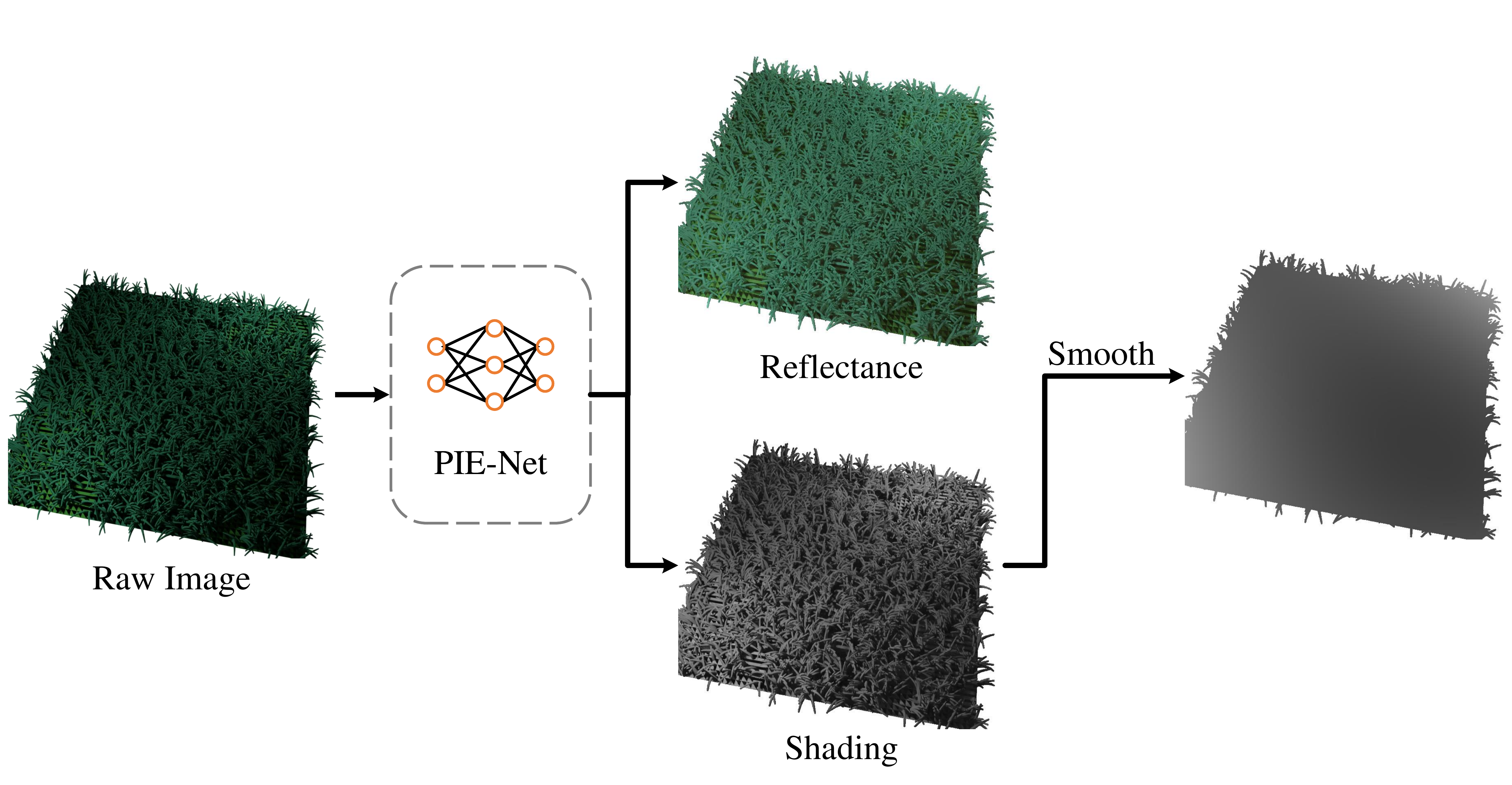}
	\caption{Preprocessing of the shading image obtained from the PIE-Net.
		\label{pre-processing}}
\end{figure}

Under natural lighting conditions, the method described above can easily synthesize new scenes of any size. 
However, when the scene lighting becomes more complex, relying solely on feature grid matching without explicit lighting guidance leads to discontinuous shading changes in the synthesized scenes, as depicted in Fig.~\ref{shadeton_compare}. 
To address this issue, we introduce the shading vector, which describes the lighting information of a basic unit in the scene. 
This vector represents the shading values of a point in the scene from different viewpoints. By incorporating the shading vector, we can ensure continuous lighting in each area of the synthesized scene, resulting in a visually coherent lighting across the entire scene.

To incorporate the shading vector into the NeRF synthesis task, we introduce two simplifications. Firstly, since the voxel column serves as the fundamental unit of NeRF synthesis, we simplify the shading vector to represent the shading values of a voxel column from different viewpoints. Secondly, accurately fitting the shading values of a point from numerous viewpoints requires substantial computational resources. Therefore, we only approximate shading vector by sampling from $M$ viewpoints. This approximation is reasonable as the primary role of the shading vector is to differentiate the brightness information among voxel columns. Through experimentation, we have determined that utilizing 50 viewpoints is sufficient for most scenes. By collecting shading information of a voxel column from multiple viewpoints within the scene, we can utilize the shading maps of the source and target scenes to synthesize a new scene that effectively captures the desired lighting effect.

To acquire the shading vectors, we project the shading value of each pixel in the training images onto the voxel grid based on depth. In this study, we utilize PIE-Net~\cite{das2022pie} as the method for intrinsic decomposition to derive the shading and albedo of the images. However, as shown in Fig.~\ref{pre-processing}, the shading obtained from PIE-Net still retains geometric information, which can interfere with the matching of shading vectors. To tackle this challenge, we apply Gaussian blurring and polynomial fitting techniques to smooth the shading image. Subsequently, we project the shading from each pixel in the smoothed shading image along the camera ray to the corresponding depth position. The depth value is obtained from the density grid using the following formula:

\begin{equation}
	d = \sum_k T_i\left(1-\exp \left(-\sigma_i \delta_i\right)\right) t_i.
	\label{eq_depth}
\end{equation}
The shading values of each voxel column in the scene from multiple viewpoints can be obtained by projecting the 2D shading information from $M$ views into 3D voxel grid. The vector formed by these values in sequential order is referred to as the shading vector, which encodes the lighting information of a voxel column. However, due to occlusions between voxels and the limited number of pixels in an image, certain voxel columns may lack shading in certain viewpoints. As a result, holes can appear in each channel image of the shading vector. And inaccurate depth values can introduce noisy points in the channel image. To address this issue, we employ the k-nearest neighbors (KNN) method to interpolate these holes, and we apply inverse distance weighting to weight the nearby points. Additionally, for noisy points, we use median filtering to repair the channel images of the shading map.
If a particular channel exhibits a high number of missing shading values, the interpolation error will increase. Therefore, we only retained the $n_c$ channels with the lowest rate of missing values.

\begin{figure}
	\centering
	\includegraphics[width=0.6\linewidth]{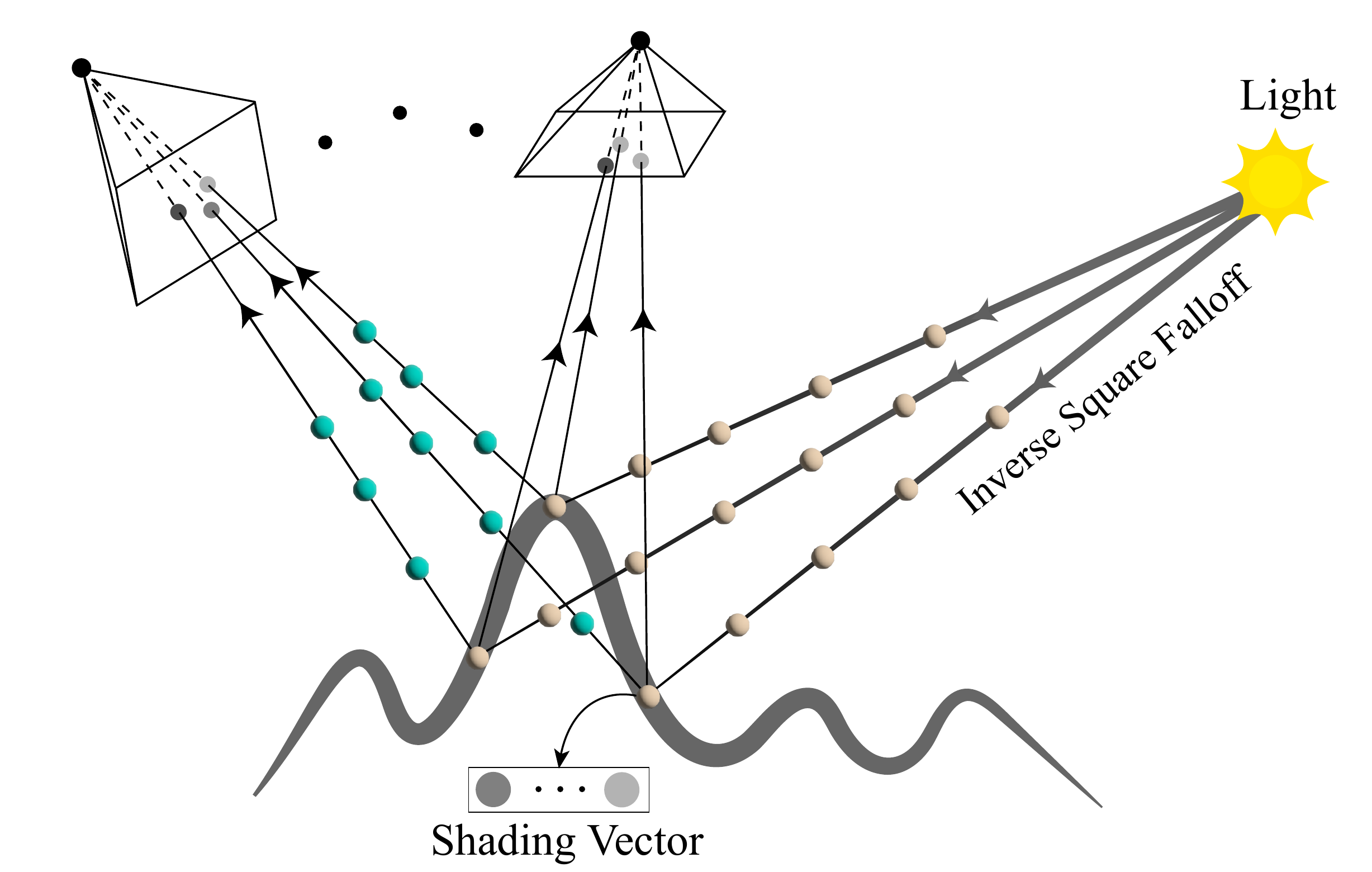}
	\caption{Shading vector from ray tracing.
		\label{shadeton_density}}
\end{figure}

Extracting the shading vector of a scene using intrinsic decomposition requires images from multiple views of the scene. However, this poses a limitation as it becomes impossible to synthesize a new scene with arbitrary lighting effects if multiple views are not available. To overcome this constraint, we propose a ray-tracing-based approach to obtain the shading vector in radiance field. This approach allows us to determine the shading information of the scene for any lighting condition by accurately tracking the path of light rays.

Specifically, the problem can be formulated as determining the shading value for each position in the scene from $n_c$ viewpoints, using the NeRF representation of the scene and information regarding the light source. The shading of a point is determined by both the intensity of its reflected light and the probability of capturing its emitted light with the camera. Additionally, the brightness of a point is influenced by the brightness of the light source and the probability of the emitted light from the source reaching the point.

Therefore, we begin by selecting $n_p$ points from the scene with a high density. Subsequently, we calculate the probability of a light ray intersecting each of these points originating from the light source. To illustrate this, let's consider a point light source, as depicted in Fig.~\ref{shadeton_density}. For any point $\mathbf{x}$ in space, we sample $n_s$ points $\mathbf{x}_1, \dots, \mathbf{x}_{n_s}$ along the path between the light source $\mathbf{x}_l$ and $\mathbf{x}$, where $\mathbf{x}_{n_s} = \mathbf{x}$. The brightness of a point can be represented by the following equation:

\begin{equation}
	b = p_{n_s}\frac{1}{||\mathbf{x}-\mathbf{x}_l||_2^2}I.
\end{equation}
Here, $p_{n_s} = T_{n_s}(1-exp(-\sigma_{n_s}\delta_{n_s}))$ represents the probability of the ray propagating to a point $\mathbf{x}_{n_s}$. The term $\frac{1}{||\mathbf{x}-\mathbf{x}_l||_2}$ corresponds to the light attenuation, and $I$ denotes the intensity of the light source. Consequently, we can determine the shading values for these $n_s$ points in the scene.
Additionally, we leverage the reversibility of light paths to convert the probability of a point's brightness reaching the camera into the probability $q_{n_s}$ of a camera ray intersecting with that point. By employing this method, we can calculate the shading of $\mathbf{x}$ from a specific viewpoint using the following equation:

\begin{equation}
	s = q_{n_s}b,
\end{equation}
where $q_{n_s} = T_{n_s}(1-exp(-\sigma_{n_s}\delta_{n_s}))$. To compute $q_{n_s}$, we apply the same sampling procedure from the camera origin to $\mathbf{x}$, as described in the previous step.
We utilize the same $n_c$ viewpoints with the lowest rate of missing values to obtain the shading vector in the specific order. The $n_s$ sampling points are obtained by projecting pixels from these viewpoints along the camera rays. The projection depth is also determined using formula~\ref{eq_depth}. To address the presence of holes in the shading map caused by occlusions and the presence of the residual of geometry details, we apply the same post-processing method as described in the previous section.


The pipeline for guiding the synthesized result with shading is shown in Fig.~\ref{pipeline}.
First, the shading map of the exemplar is used to define the multi-view shading of each voxel column, while the shading map guider is used to specify the lighting information of the target scene.
The shading map is obtained using the intrinsic decomposition method, as multiple images of the exemplar are available.
To obtain the shading map guider, we synthesize scene using geometry-based synthesis to create a geometrically continuous large scene with larger size (without appearance-based method, we directly find the patch with the closest density at the overlap). Then, the ray-tracing-based method is employed to generate the shading map guider for the target illumination.
For the patch $P_n$ to be synthesized, the L2 distance is initially utilized to identify the $k_s$ shading patches in the shading map that are closest to the shading of $P_n$, which is accelerated by the kd-tree.
As the shading map and shading map guider are obtained from different methods, it is necessary to normalize the values of them to the same scale. However, the selection of patches can be limited due to the requirements of shading, geometry, and texture. To address this limitation, we rotate the exemplar by 90, 180, and 270 degrees, respectively, to obtain additional available patches. Finally, a two-phase method is employed to select patches that satisfy both the geometric and texture continuity requirements from these $k_s$ patches.

How to attach realistic surface structures onto mesh surface has been a long-standing problem, as the quality of modeling directly affects the visual effect. With the aid of shading vector, we can synthesize scenes with arbitrary lighting effects on 3D mesh and apply the surface structure modeled by NeRFs to achieve mixed rendering of the two representation methods.

The attachment of realistic surface structures onto a mesh surface has long been a challenging problem, as the quality of modeling directly impacts the visual outcome. By leveraging shading, we can synthesize scenes with various lighting effects on a 3D mesh and incorporate the surface structure modeled by NeRFs.

To achieving this goal, first, we synthesize a scene on a plane with the desired lighting using a specific shading map guider. Then, we deform the corresponding voxel grid to match the 3D mesh using a deformation field.
To obtain the shading map guider, we utilize the deformation field to bend the synthesized large-scale $V^{(d)}$ onto the 3D mesh. Subsequently, we employ the ray-tracing-based method to compute the shading map guider on the surface.
Finally, we utlize the shading map guider to synthesize a new scene with desired  lighting and deform it to match the target mesh surface.
The deformation field is represented as $\Psi: \mathbf{x} \rightarrow \Delta \mathbf{x}$, which maps points in the deformed space to the canonical space. To obtain the deformation field, we employ the ARAP parametric method~\cite{liu2008local} to establish the mapping between the 3D mesh and the undersurface of the scene.
Furthermore, we define the corresponding direction for each vertex on the mesh and its corresponding point on the plane along the normal direction. This correspondence allows us to collect a large number of discrete corresponding points in the deformation field. We employ a neural network to fit this deformation field and train the network $\Psi$ using the coordinates of these corresponding points as supervision information.

\section{Experiments}

\subsection{Datasets}
We evaluated the effectiveness of our method mainly on synthetic data. For each synthetic scene, we assume the volume of interest is within a unit cube, and 100 observation images fully covering the scene are collected with cameras randomly distributed on a hemisphere. 

To ensure the generalization ability of our method in real-world scenes, we also collected a significant amount of real-world data and tested our approach on them. 
Given the need for high geometric accuracy in NeRF synthesis and the inherent complexity of real-world scenes, we employed 300 images for real-world scenes. To eliminate background interference, we manually removed the background information from each image. Additionally, for each scene, we calibrated the images using colmap~\cite{schonberger2016structure} to determine the pose of them.

\subsection{Implementation details}
We choose the same hyperparameters generally for all scenes. All hyperparameters of DVGO are consistent with those in their article.
In particular, we initialize the shape of grid to $160\times 160 \times 160$.
The patch width and overlap area is set to $w_{p}=15$ and $w_{o}=5$ respectively. To generate the template patch within the original grid, we utilized a sliding window with a step size of 3.
For the patch search process, we utilized $k_s=20$ and $k_g=10$ to locate suitable patches.
Determining the length of the boundary was performed individually for each sample, ensuring that the boundary region of the grid encompassed all the physical boundaries.

For preprocessing the shading map, we applied a 10-times Gaussian blur with a mean of 0 and a variance of 7. Additionally, we performed cubic polynomial fitting on all channels of the shading map, which was obtained using two different methods, for all scenes.

We use a fully-connected network architecture to represent the deformation fields $\Psi$. The input position is passed through 3 fully-connected layers with ReLU activations, each with 16 channels.
We utilize the MSE loss function to estimate the loss and employ the Adam~\cite{kingma2014adam} optimizer to optimize the network. For the Adam optimizer, we keep all the parameters at their default values.

\subsection{Ablation}

\begin{figure}
	\centering
	\includegraphics[width=0.8\linewidth]{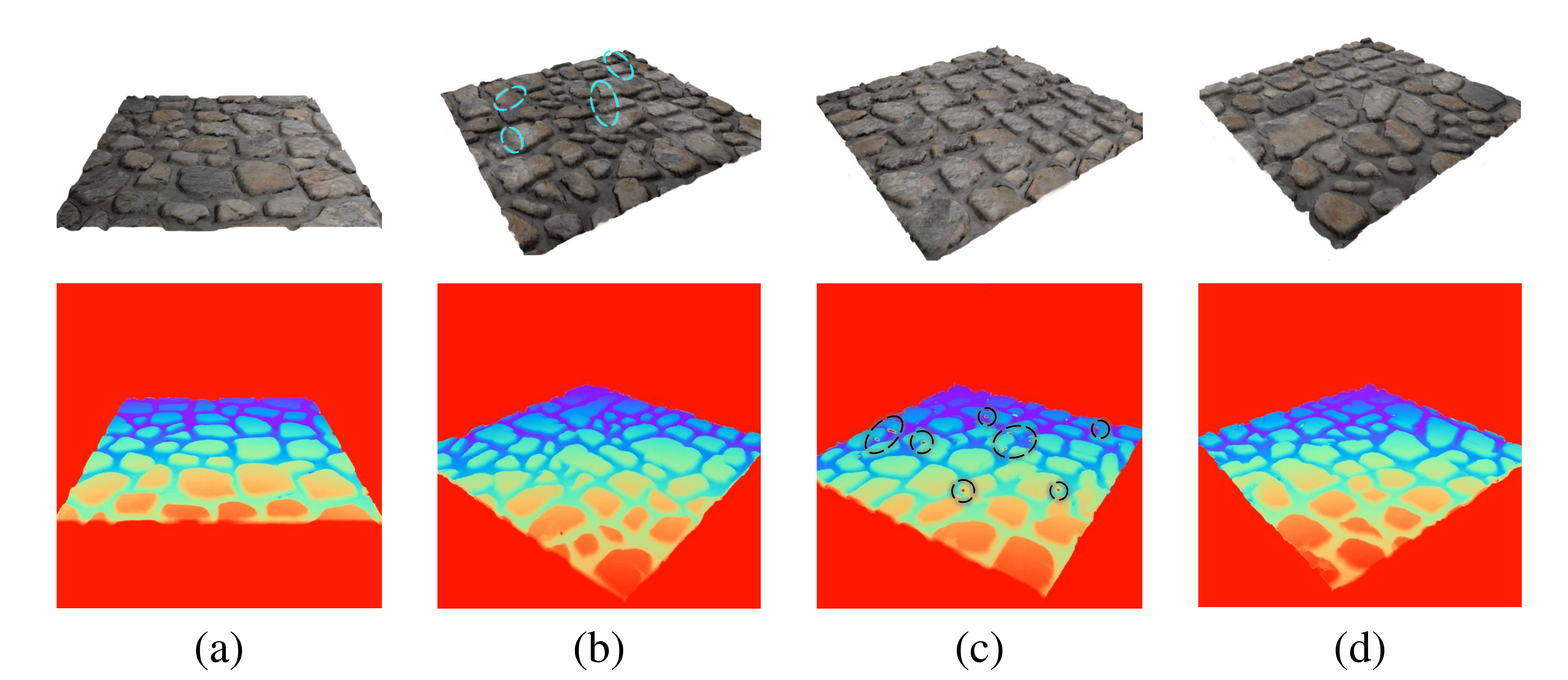}
	\caption{(a) Exemplar: rock~\cite{rock}. RGB and depth images are shown.
		(b) Synthesis based on higher density weights. Clear color discontinuities can be observed in the RGB image (blue circles).
		(c) Synthesis based on higher feature weights. Geometric discontinuities can be observed in the depth image (black circles).
		(d) Results of the two-phase method. Both appearance and geometry exhibit continuity similar to the exemplar.
		\label{two-phase}}
\end{figure}

\begin{table}[t]%
	\caption{Comparison of the two-phase method with the baseline in terms of speed when synthesizing results of different sizes. The experiment does not include the part of shading guided synthesis.\label{tab:speed}}%
	\centering
	\begin{tabular}{@{\extracolsep\fill}lcccccc@{\extracolsep\fill}}
		\hline
		Size & $100^2$  & $200^2$  & $300^2$  & $400^2$\\
		\hline
		Baseline (s) & 163.47 & 270.70 & 454.41 & 714.64\\
		Two-phase (s) & 14.99 & 23.62 & 38.46 & 59.57\\
		Times & 10.90 & 11.46 & 11.82 & 12.00\\
		\hline
	\end{tabular}
\end{table}

The exemplar and the results obtained from different synthesis methods are depicted in Fig.~\ref{two-phase}. It demonstrates that our proposed two-phase method effectively achieves a balance between appearance and geometric continuity. Furthermore, the two-phase method significantly enhances the synthesis speed by more than 10 times compared to the baseline method, as demonstrated in Table~\ref{tab:speed}, since we only need to search patches for density in the first phase. 

\begin{figure}
	\centering
	\includegraphics[width=0.6\linewidth]{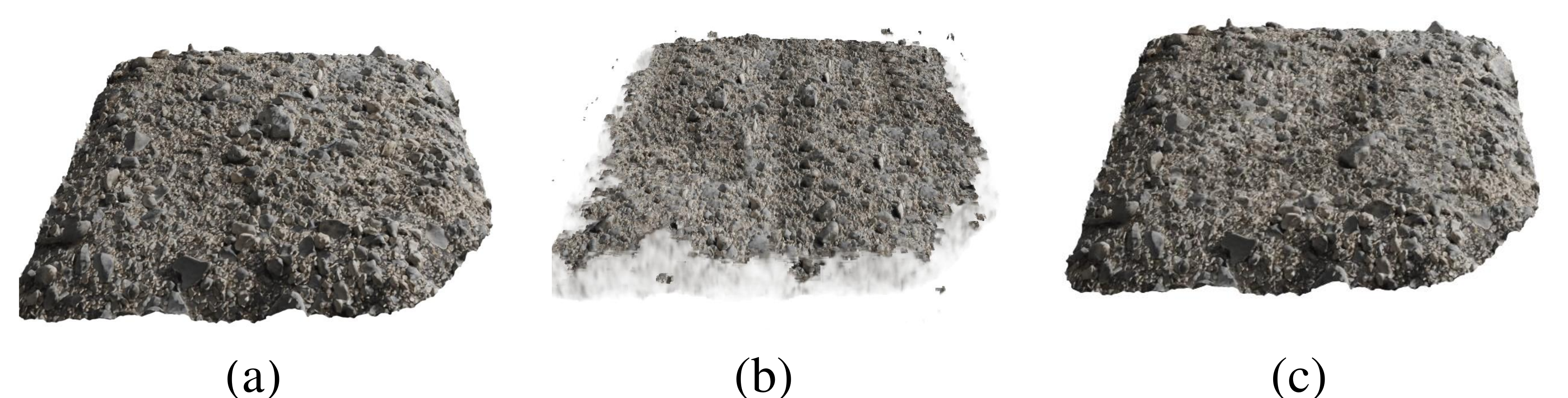}
	\caption{(a) Exemplar: ground~\cite{ground}.
		(b) Synthesized result without boundary constraint: severe artifacts are observed at the boundaries.
		(c) Boundary-constrained synthesis method: the synthesized result is nearly indistinguishable from the exemplar.
		\label{boundary}}
\end{figure}

Fig.~\ref{boundary} illustrates that artifacts are generated at the boundary using the two-phase method. However, by incorporating the boundary constraint, we can obtain superior results at the boundary.

\begin{figure}
	\centering
	\includegraphics[width=0.6\linewidth]{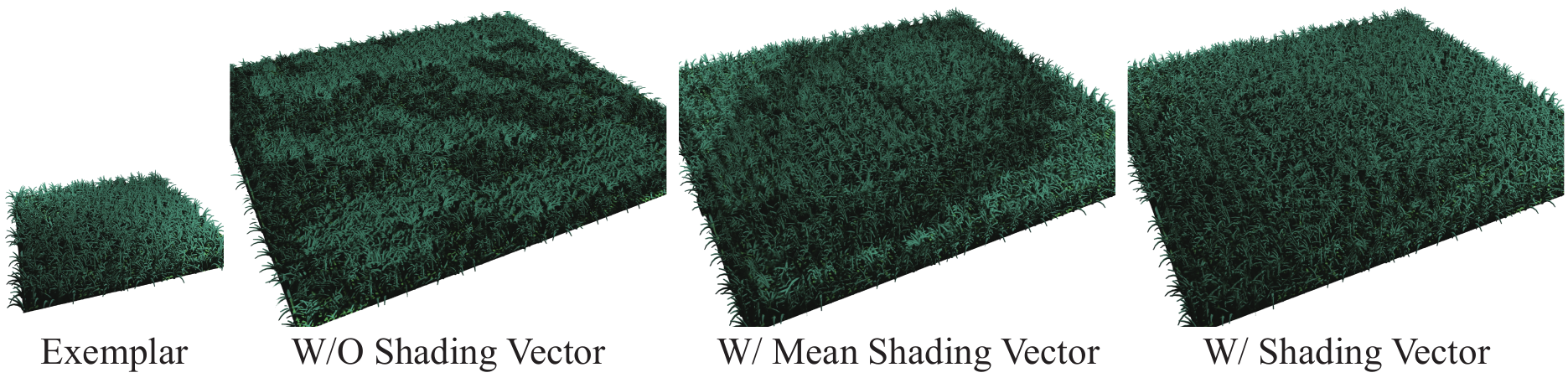}
	\caption{Synthesized results of a grass~\cite{grass} with complex lighting. Such scenes cannot be handled without shading vectors or with mean shading vectors. The shading vector can be used to synthesize results with continuous illumination. The shading map guider used in this experiment is obtained by doubling the size of the shading map from the exemplar.
		\label{shadeton_compare}}
\end{figure}

To demonstrate the effectiveness of the shading vector, we performed an averaging operation on all channels of the shading vector, removing the luminance distribution of the shading vector across multiple views. We then used only the average value to describe the lighting information of a voxel. Fig.~\ref{shadeton_compare} shows that the average shading alone does not ensure luminance continuity in the synthesized scene from each viewpoint, whereas the shading vector produces better results.

\subsection{Results}

\begin{figure*}
	\centering
	\includegraphics[width=\textwidth]{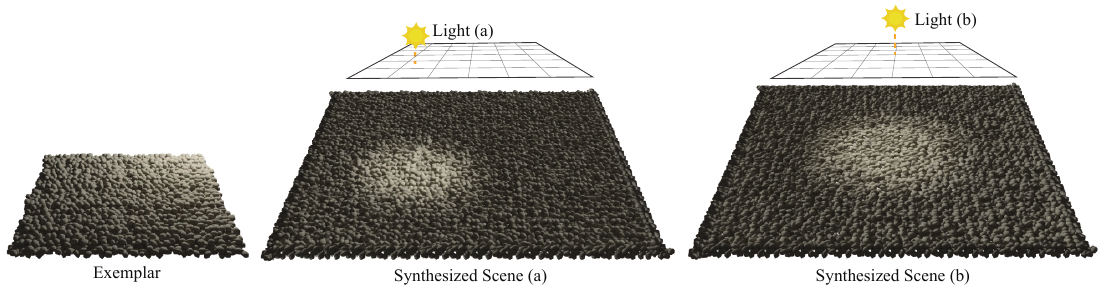}
	\caption{Relighting the scene of pebbles~\cite{pebbles} using different lighting conditions.
		\label{results_relighting_2}}
\end{figure*}

\begin{figure*}
	\centering
	\includegraphics[width=\textwidth]{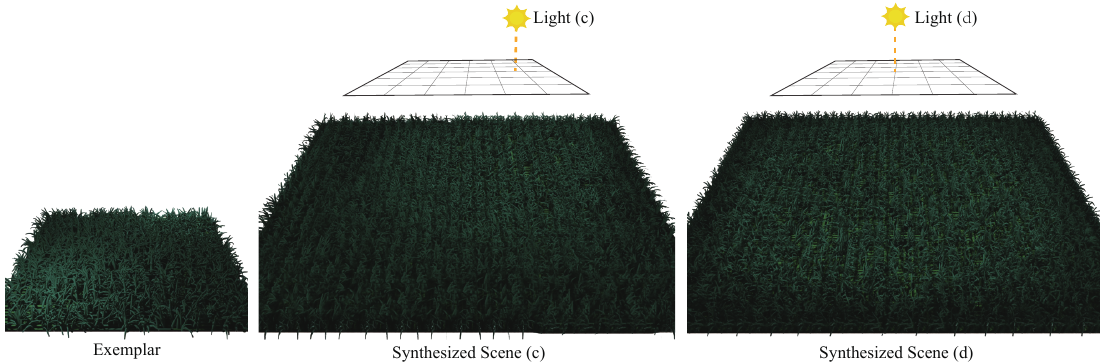}
	\caption{Relighting the scene of grass using different lighting conditions.
		\label{results_relighting_3}}
\end{figure*}

\begin{figure*}
	\centering
	\includegraphics[width=\textwidth]{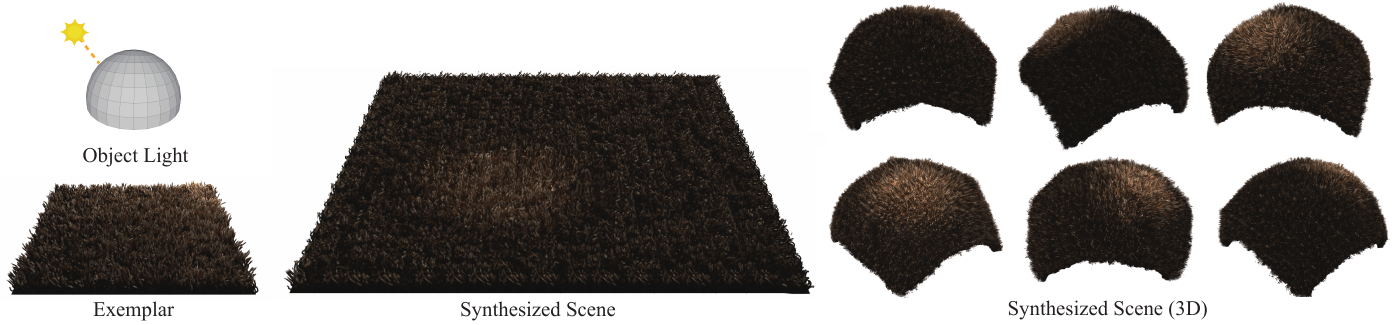}
	\caption{Relighting the scene of fur~\cite{couch} on a 3D sphere. The synthesized scene on the plane is obtained by using the shading map extracted from the scene bent onto the surface of the 3D sphere as a guide for synthesis. The 3D synthesized scene is achieved by bending the synthesized scene on the plane onto the sphere.
		\label{3d_ball}}
\end{figure*}

There are two approaches to synthesize new scenes with continuous illumination effects, differing only in the choice of the shading map guider. 

The first approach obtains the shading map guider using a ray-tracing-based method. This method allows for the generation of new scenes with different lighting effects and varying sizes compared to the exemplar, as depicted in Fig.~\ref{results_relighting_2} and Fig.~\ref{results_relighting_3}.

To synthesize a new scene with specified lighting effects on a curved surface, we synthesized a new shading map guider on a sphere using the ray-tracing-based method and then employed it on a plane to guide the synthesis of a large-scale scene with the desired lighting effects. The synthesized result was subsequently mapped onto the sphere using the deformation field, as illustrated in Fig.~\ref{3d_ball}. 

\begin{figure*}
	\centering
	\includegraphics[width=\textwidth]{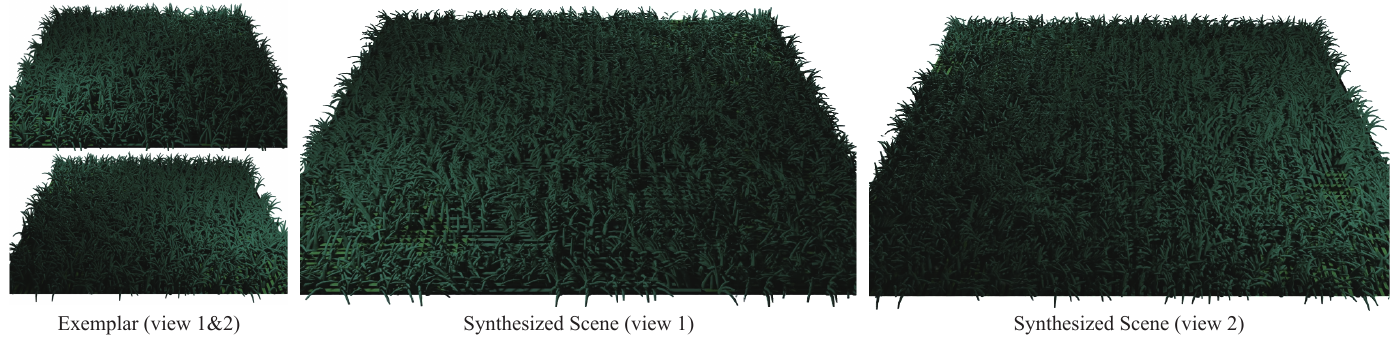}
	\caption{The synthesized results of maintaining the lighting patterns. The shading map guider used in this experiment is obtained by doubling the size of the shading map from the exemplar.
		\label{results_relighting}}
\end{figure*}

The second approach to obtaining the shading map guider involves directly scaling up the shading map of the exemplar. This method results in a new large-scale scene with the same lighting pattern as the exemplar, as demonstrated in Fig.~\ref{results_relighting}.

\section{Discussion}
This article introduces a noval task called NeRF synthesis. It requires synthesizing a large-scale scene using a small exemplar represented by NeRF, while ensuring the continuity in terms of geometry and appearance of the synthesized scene.
Applying traditional patch-based synthesis methods directly to the NeRF synthesis problem will result in some new issues. Firstly, in order to balance the weights of geometry and appearance, and address the boundary problem in NeRF synthesis, we propose a two-phase method and a boundary-constrained synthesis method, respectively. We evaluate the effectiveness of these methods on diverse scenes with complex geometry and rich textures, demonstrating their ability to generate high-quality synthesized results. Secondly, due to the coupling of lighting and material in NeRF, the lack of explicit lighting guidance can result in discontinuous lighting effects in the synthesized results. To overcome this, we propose the use of shading vectors to guide the synthesis process by representing the lighting information for each unit in the scene. For scene relighting, we develop two approaches for obtaining shading vectors and demonstrate through experiments that their combination can effectively achieve relighting without completely decoupling the scene. Our proposed methods provide a novel approach to synthesizing 3D objects, particularly well-suited for synthesizing natural environments such as lawns, ground, and stone surfaces in virtual worlds.

\textbf{Limitations}: Our method represents the scene directly as a grid, but scaling up the scene leads to higher memory consumption. To achieve optimal performance, our proposed approach requires a high-quality DVGO as input, which often requires a larger number of input images for real-world scenes. Since the relighting task requires continuity in geometry, texture, and lighting simultaneously, it is not suitable for scenes with complex geometry changes, as finding a suitable patch in the exemplar may not be possible. Furthermore, as capturing scenes with complex lighting effects in real-world settings is very challenging, we still lack experiments with re-lighting in real-world scenes.

\textbf{Future work}: Our future research plans involve exploring the application of learning-based approaches, such as GAN-based and diffusion model-based methods, to tackle NeRF synthesis problems. Additionally, although we have achieved successful scene synthesis on 3D meshes, these experiments have been limited to simple mesh structures. Therefore, our aim is to investigate how NeRF synthesis can be performed on arbitrary curved surfaces, which presents a substantial challenge and an opportunity for further research. This exploration would enable the seamless integration of NeRF with traditional rendering techniques.

\bibliographystyle{unsrt}  
\bibliography{references}

\section*{Supplementary Material}
\begin{figure*}
	\centering
	\includegraphics[width=\textwidth]{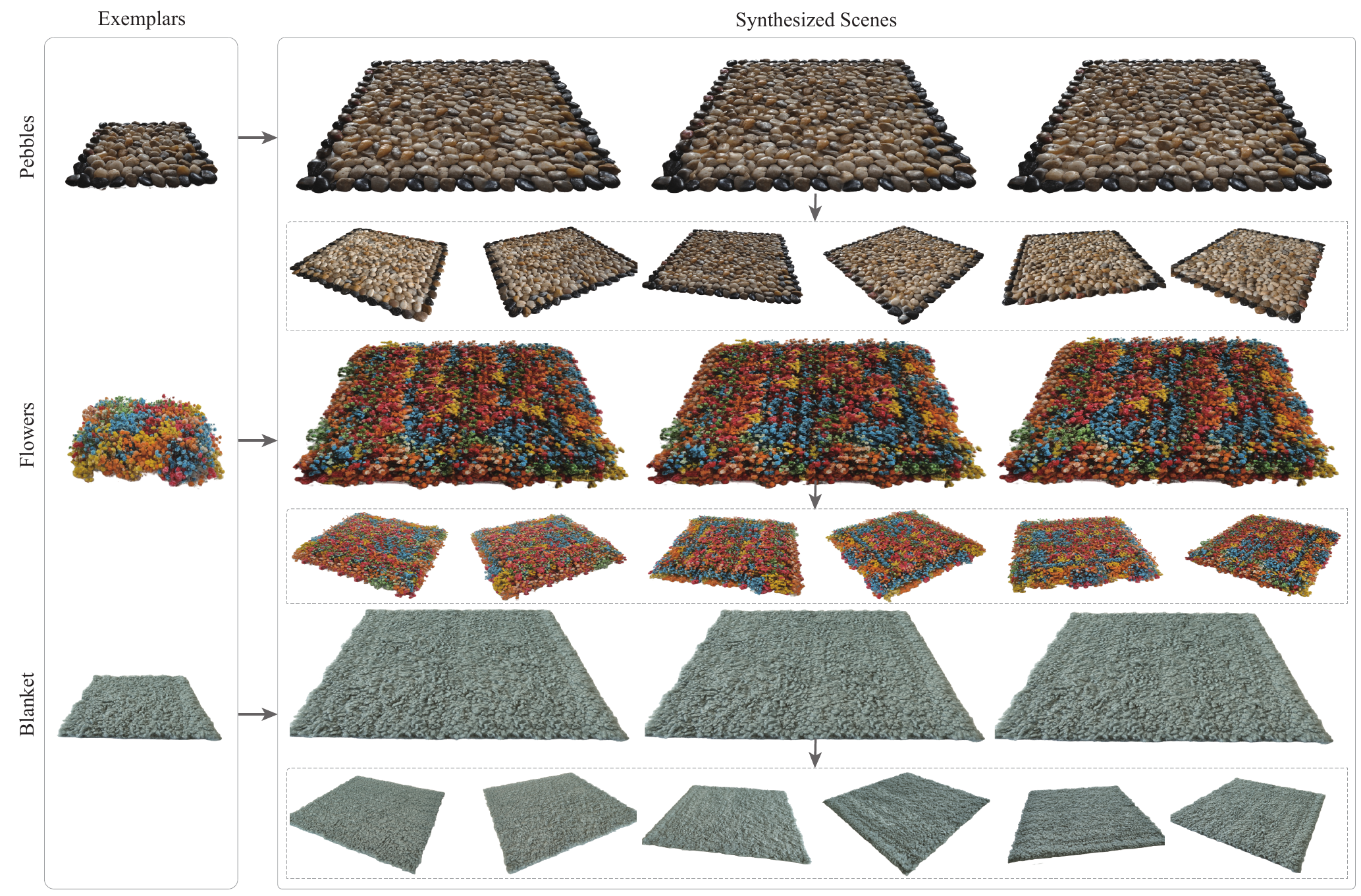}
	\caption{Synthesized results of real-world scenes with natural light. The odd rows display the various synthesized results, while the even rows show multi-view images of one of the synthesized results.
		\label{results_real}}
\end{figure*}

\begin{figure*}
	\centering
	\includegraphics[width=\textwidth]{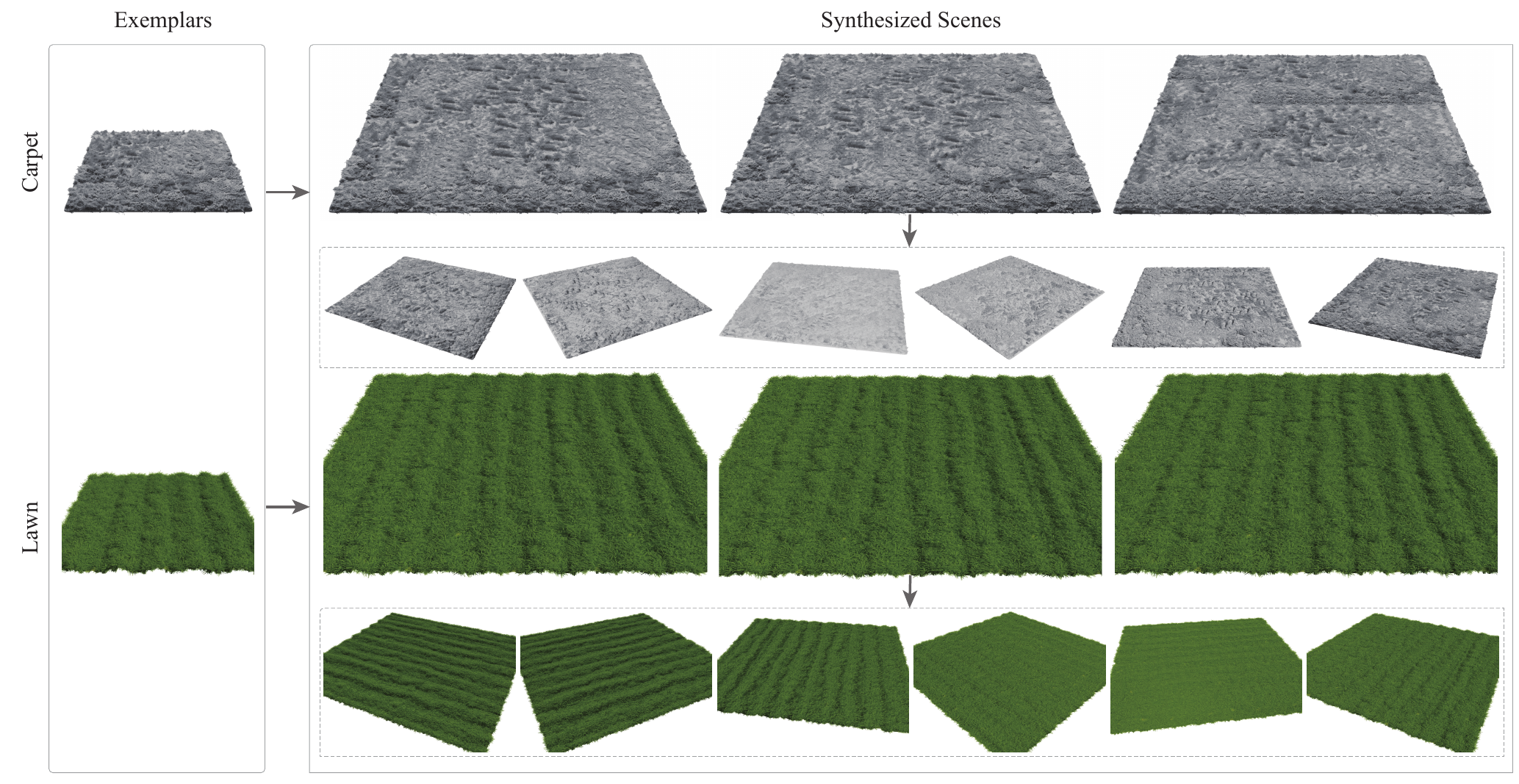}
	\caption{We present synthesized results under natural lighting conditions, using carpet~\cite{carpet} and lawn~\cite{lawn} as exemplars. The odd rows display the various synthesized results, while the even rows show multi-view images of one of the synthesized results.
		\label{results_synthetic}}
\end{figure*}

For natural light scenes, the consideration of lighting effects is not necessary, thus shading is not required to guide the NeRF synthesis process.
We conducted extensive testing of our method on challenging synthetic and real-world data with complex geometry and vibrant colors under natural lighting conditions. The results, as illustrated in Fig.~\ref{results_synthetic} and Fig.~\ref{results_real}, demonstrate that our method achieves synthesized results that are nearly indistinguishable from the exemplar. It is important to note that our method is not limited to scenes with distinct geometry boundaries like rocks and ground, but can also handle scenes with intricate geometry such as grass and lawns.

\begin{figure*}
	\centering
	\includegraphics[width=\textwidth]{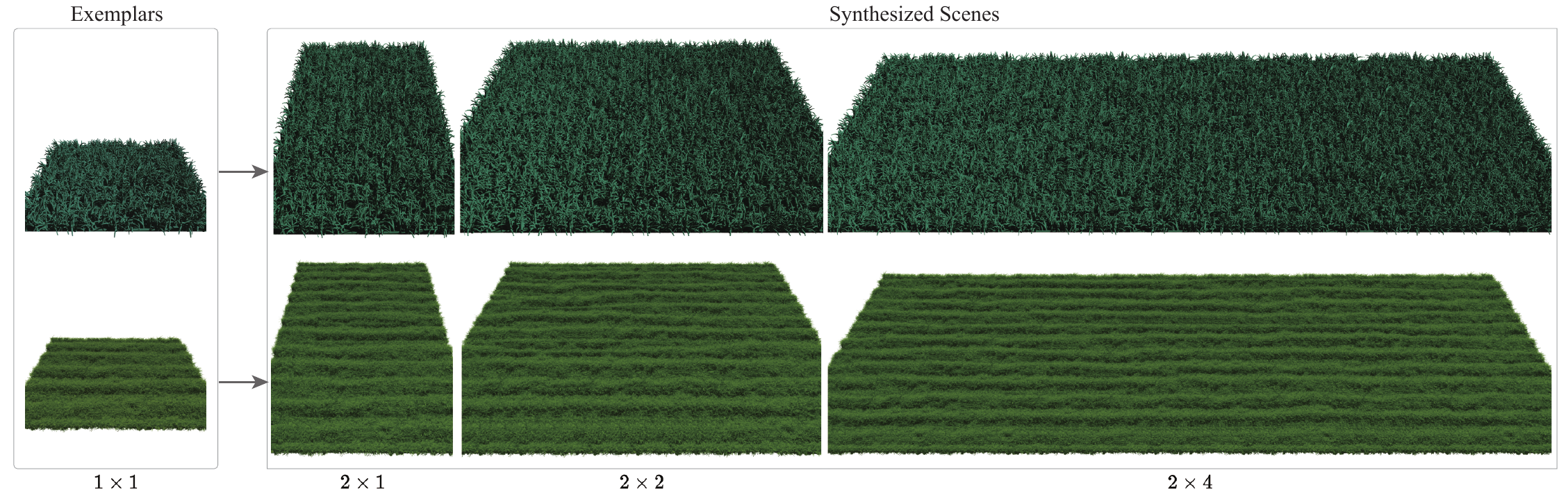}
	\caption{Synthesized results for larger sizes with natural light.
		\label{results_scale}}
\end{figure*}

Due to space limitations, only the results of synthesizing the exemplar at double the size are presented in the main text. However, it is worth mentioning that our method is capable of synthesizing results of any size and aspect ratio, as exemplified in Fig.~\ref{results_scale}.


\end{document}